\pdfoutput=1
\documentclass[11pt,a4paper]{article}
\usepackage[]{acl}

\usepackage{times}
\usepackage{latexsym}

\usepackage[T1]{fontenc}
\usepackage[utf8]{inputenc}

\usepackage{algorithm}
\usepackage{booktabs}
\usepackage[noend]{algpseudocode}

\usepackage{graphicx}

\usepackage{microtype}

\newcommand\Islander{Islander}

\title{\Islander: A Real-Time News Monitoring and Analysis System}

\author{Chao-Wei Huang$^{\dag,1}$, Kai-Chou Yang, Zi-Yuan Chen, Hao-Chien Cheng, Po-Yu Wu,\\
\bf Yu-Yang Huang, Chung-Kai Hsieh, Geng-Zhi Wildsky Fann, Ting-Yin Cheng,\\
\bf Ethan Tu, and Yun-Nung Chen$^{\star,1}$ \\
Taiwan AI Labs, Taipei, Taiwan \\
$^{1}$National Taiwan University, Taipei, Taiwan \\
\texttt{$^\dag$f07922069@csie.ntu.edu.tw \quad $^\star$y.v.chen@ieee.org %\{zake7749,zychen,payo,wupoyu,yuyang.huang,}
}
}

\begin{document}
\maketitle

\begin{abstract}
With thousands of news articles from hundreds of sources distributing and sharing everyday, news consumption and information acquisition have been increasingly difficult for readers.
Additionally, the content of news articles are becoming catchy or even inciting to attract readership, harming the accuracy of news reporting.
We present \Islander, an online news analyzing system for online news. % in Taiwan
The system allows users to browse trending topics with articles from multiple sources and perspectives.
We define several metrics as proxies to news quality, and develop algorithms for automatic estimation.
The quality estimation results are delivered through a web interface to news readers for easy access to news and information~\footnote{The website is publicly available at https://islander.cc/}
% \footnote{Demo video at https://youtu.be/aKOa-q7FqYs}

\end{abstract}

\section{Introduction}
News consumption has shifted to online media in the past two decades.
According to Reuters Institute Digital News Report~\cite{newman2015reuters}, about 75\% users use online media as a news source at least once a week.
At the same time, media outlets have also been putting more effort into online media to accommodate the growing demands.
Going online has brought several benefits to the publishers, e.g., improved accessibility, easy distribution, reduced paper waste, time constraint-free, and rapid spreading.
With the strong capabilities of search engines and the fast-growing social platforms, news dissemination has been exponentially faster in the internet era. 

Due to the for-profit nature of most online media outlets, the most important measurement of success is popularity.
While the algorithms of search engines and social platforms provide an efficient channel for information diffusion, it is observed that the \emph{quality} and the \emph{accuracy} of content are not the only factors for news to be popular.
A study has shown that algorithms that put too much weight on popularity can hinder quality~\cite{ciampaglia2018algorithmic}.
\citeauthor{pennycook2021shifting} showed that when deciding what to share on social media, users are often distracted from considering the accuracy and the veracity of the content.
Instead, users tend to share the news that attracts viewers' attention the most.
Therefore, a misalignment between quality of content and profitability is created.
As a result, online media outlets often publish pieces with clickbait headlines or exaggerating content to increase readership~\cite{blom2015click}.
The user engagement is not generated through quality, but rather through catchy, provocative headlines, making the news to be subjective, biased, misleading, or even worse, inaccurate.
Several studies have researched the application of natural language processing (NLP) technologies for detection of such behavior~\cite{chakraborty2016stop,potthast2016clickbait,potthast2018crowdsourcing}.
However, most previous work focused on the detection and analysis of clickbait headlines without studying the effect of content.

Media with a large user base is highly-influential to readers' perception of events, directly~\cite{mcleod1992manufacture} or indirectly~\cite{gunther1998persuasive,gunther2003influence}.
Therefore, such media outlets are capable of steering public opinion towards a desired direction.
It is shown that news media could change opinion toward a presidential candidate by varying news coverage during presidential campaign~\cite{morris2010cable}.
Even worse, news media could spread misinformation which results in stronger public opinion manipulation~\cite{guess2018selective}.
In the mean time, online forums also play an important role in misinformation diffusion.
Readers usually post news to online forums for discussion.
However, such discussion is often prone to manipulation by a few opinion leaders, making it a target for public opinion manipulation.

In this paper, we present \Islander, an online news analyzing system, where the goals are:
1) identifying news that are exaggerating, subjective or inaccurate,
2) identifying public discussions of news that are possibly manipulated, and
3) allowing users to efficiently explore high-quality content.
We develop crawlers to retrieve news reports from online media outlets, and build models to analyze the quality of the news articles based on the following properties:
\begin{enumerate}
    \item {\bf Objectivity}: news reports should be delivered with factual descriptions, not subjective judgments from the reporters.
    \item {\bf Neutrality}: news reports should be unbiased, describing the incidents without a personal stand.
    \item {\bf Non-incitement}: inciting or extreme word usage makes the piece uninforming and unprofessional. Such usage should be avoided.
\end{enumerate}

The system comes with a web interface that provides several functionality for users:
1) browsing trending topics in a chosen time period,
2) comparing news metrics for different online news media, and
3) providing charts to show the news metrics and important news in a chosen time period.

\section{Related Work}
Previous work has been studying the quality of news reports in the online media landscape.
\citeauthor{ramirez2015differences} compared digital and printed version of five European newspapers, and confirmed that there is a deterioration in news quality online.
\citeauthor{orosa2017use} offered an analysis on newspapers of the 28 EU member countries. The authors confirmed the presence of clickbait in most of the newspapers analysed.
Our system does not estimate quality or detect clickbait directly. Rather, we define desired properties as proxies to quality, and analyze news reports with the properties.

Several systems have been developed for analyzing news and information online.
Prta~\cite{da-san-martino-etal-2020-prta} categorized propaganda techniques into 18 different types, and developed a web interface that allows users to explore news articles by highlighting the propaganda techniques used in them.
ClaimPortal~\cite{majithia-etal-2019-claimportal} builds a web-based platform for monitoring, searching, checking, and analyzing English factual claims on Twitter. The design goal is to assist fact-checkers in analyzing factual claims on social media in the American political domain.
Watch ’n’ Check~\cite{9260012} builds a social media monitoring tool for collaborating with fact-checkers to detect misinformation online by assisting them in the identification and targeting of misleading claims.
NSTM~\cite{bambrick-etal-2020-nstm} provides a summarization engine that condenses large volumes of news into short, easy-to-absorb points. The engine helps users digesting the key news about companies or countries.

\begin{figure*}[t!]
\centering
\includegraphics[width=\linewidth]{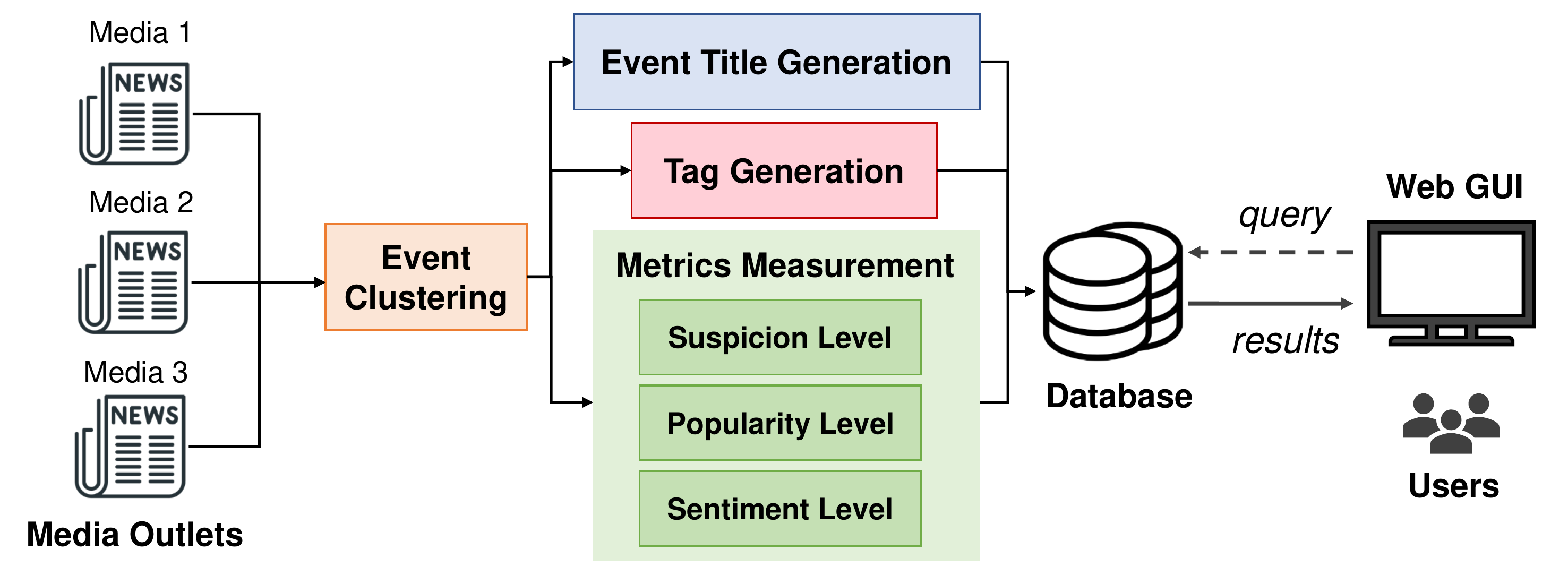}
\caption{An overview of \Islander.}
\label{fig:overview}
\end{figure*}

\section{{\Islander} System Framework}

%\subsection{Overview}
An overview of our system architecture is illustrated in Figure~\ref{fig:overview}.
We collect news articles from multiple media sources.
Considering that many articles may describe the same events, digesting news information should be event-based for better efficiency and completeness.
Hence, the collected articles are grouped with a proposed event clustering module, and a tag generator focuses on generating associated tags for describing the events.
These components are detailed as follows.
After clustering, we apply a metrics measurement module for both news articles and their associated public opinions (detailed in Section~\ref{sec:qual}); then the results are stored in a database.
Finally, the users can interact with the web interface to input queries and get the results shown on the webpage.

\subsection{Real-Time News Crawler}
Based on the popularity of the online news media outlets, we select 24 online news media to track.
We develop crawlers for each news media.
The crawlers are checking new articles regularly to ensure the database is up-to-date.

\subsection{Event Clustering}
A special property of our system is that we process and analyze news in an event basis, rather than article basis.
After we collect the news articles, we will first perform event clustering to group articles into event clusters.

The event clustering module consumes news articles with a per-batch procedure, where articles focusing on similar topics with possibly different perspectives that are published within a short period of time are processed to form event clusters.

\subsubsection{Batch Clustering}
\label{section:batch_clustering}
In order to create event clusters, we process articles within a short period of time in batches. To focus on most broadcasted news event, we develop a new clustering method, as shown in Algorithm~\ref{alg:clustering}.

\begin{algorithm}[t]
\caption{Batch Clustering}\label{alg:clustering}
\begin{algorithmic}[1]
\State $t_1, t_2 \in R $

\State $D = \{d_1, d_2, ...d_n\}$
\State $C_i = \{\forall d_j \in D: similarity(d_i, d_j) < t_1\}$
\State $f(C_i, C_j) = \frac{|C_i \cap C_j|}{\min(|C_i|,|C_j|)}$

\State $S = \{C_1, C_2, ... C_n\}$
\While{ $\exists C_i, C_j$ such that $f(C_i, C_j) > t_2$}
\State $C_k = C_i \cup C_j$\Comment{Merge cluster pair}
\State remove $C_i, C_j$ from $S$
\State add $C_k$ into $S$

\EndWhile\label{euclidendwhile}
\State \textbf{return} $S$
\end{algorithmic}
\end{algorithm}

For each article $d_i$, we create an initial cluster $C_i$, which contains other articles that has cosine similarity higher than a given threshold $t_1$ in the latent space.
The initial clusters may overlap, so we merge a pair of clusters if their overlapping ratio, $f(C_i, C_j)$, is higher than a given threshold $t_2$.
The merge process will run until there is no cluster pair matching the condition.
After the merge process finishes, each article might be assigned to multiple clusters, and we will assign the largest cluster as the final label.

\subsubsection{Clustering over Time}
In order to construct clusters across days to represent an overview of a specific event, we introduce a sliding window mechanism over time.
Our system applies batch clustering with a window size of 3 days, stride step of 8 hours.
Since an article might participate in multiple clustering processes with different sliding windows, we assign an unlabeled article to a new cluster label if the cluster consists only unlabeled articles
Otherwise, the cluster label will be decided by the majority of labeled articles within the cluster. %The cluster label will be assign only one time.

\subsection{Event Title Generation}
After the events are clustered, we select a title from the articles in the cluster to represent the event.
To maintain the quality of event titles, extractive summarization is applied. 
%bstractive based model, like GPT-2, may be fancy, but it is easy to get out of control. 
We use TextRank~\cite{mihalcea2004textrank} to select a title within a group of news titles.
% TextRank algorithm is based on PageRank.
We use news titles and contents as nodes in the TextRank algorithm, but only news titles as candidates.
Therefore, the ranking is determined by titles and contents, and then we select the highest ranked title as the title of the group.
We have tried using only news titles as nodes, which resulted in worse performance.

% For evaluation, there is no ground truth and labelling for this task is costly.
% We instead compute a measure of word overlap using the ROUGE-1 score between the selected title and all candidate titles.
% We randomly sample 100 events and get 38.56 word overlap score.

\subsection{Tag Generation}
% Justin.
% - overview.
% - method.
% - experiment.

To facilitate users to understand the news event, an abstractive tag generator is proposed. The generated tags can also help optimize the retrieval performance of relevant events.
The architecture of our proposed tag generator follows the training framework of Grover~\cite{zellers2019defending}, a GPT-based language model for news generation. 

Specifically, we collected a corpus of 0.7 millions of news pages with tags written by journalists from various media sources. A news page is modeled by a joint distribution of four fields: \emph{title}, \emph{content}, \emph{news sources}, and \emph{tags}. 
During the training phase, we randomly sample a batch of fields from the corpus, and ask the model to predict the remaining fields.
For example, given the content and the source, the model will be asked to generate the titles and the tags.
To formulate the fields to texts, we wrap the content of each field with their corresponding special tokens, such as \textbf{[TITLE-START]}, and concatenate the wrapped texts. 
We generate the training examples dynamically by running the sampling process on every batch, which helps the generator learn the coherence between news fields. 

The whole training process consists of two phases.
In the first phase, we randomly sample fields as inputs and take the remainings as targets for language modeling.
While in the second phase, we gradually decrease the sampling weights of the tag field, i.e., increasing the probability of tag field being the supervised targets, in order to well-fit the downstream scenarios.
During the inference phase, the news title, content and source are given and the model is asked to generate the tag sequence.

% 這張圖的原始資料在 https://docs.google.com/spreadsheets/d/15J5auiQk9m1btLCSmcS-WOb_MhZTMVeFeL5NvLVXB1Q/edit#gid=0
% 有需要調整可以直接改

\begin{figure}[t]
\centering
  \includegraphics[width=\linewidth]{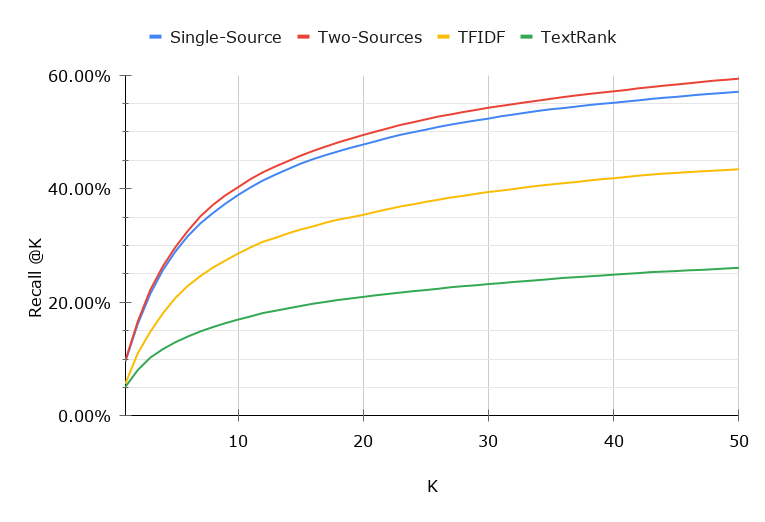}
  \caption{The performance comparison for tag generation.}
  \label{fig:tag_generator_performance}
\end{figure}

We compare our tag generator against two baseline models: TFIDF and TextRank~\cite{mihalcea2004textrank}.
The experimental result is shown in Figure~\ref{fig:tag_generator_performance}, where the evaluation metric is recall@K.
We observe that our generator significantly outperforms the baselines for all $K$, especially for a larger $K$.
We hypothesize that the reason is that about $20\%$ of tags do not appear in the news content, which is unfavorable to extractive models.
Furthermore, our keyword generator is able to generate high-level concepts, such as topics and name entities, which is much similar to how journalists would tag the news.

The performance can be further improved by simply mixing the predictions from different news sources as test-time augmentation.
In particular, for a news article, we first generate tags using the same title and content, while using different news sources.
Then, we extract the union from generated results as a robust set, and add tags with high confidence from the remainings to the robust set, until the size of robust set reaches K.
In the production stage, we only generate the tags using two news sources for efficiency.
The experimental results show that the performance can be consistently improved across all $K$ by selecting the high agreement predictions from two news sources.

\section{Metrics Measurement}
\label{sec:qual}
Our system focuses on monitoring the news information from an event basis, so we define three metrics: \emph{suspicion}, \emph{popularity}, and \emph{sentiment} levels, in order to better show the worth-noting news events in the interface.

%\subsection{ Metrics for News Articles}
% suspicious -> incite of title, subjectivity of content, stance of title
% popularity -> num of news and num of related comment
% sentiment -> sentiment of news content & sentiment of comment

%On \Islander, we calculate the degree of \emph{suspicion}, \emph{popularity}, and \emph{sentiment} of each news.

\subsection{Suspicion Level in News}
We determine the degree of suspicion by the following factors: incite of title, bias of title, and subjectivity of content.

\paragraph{Incitation of the title} is calculated as the average inciting score within a group of news titles. 
CVAT data~\cite{yu2016building} contains the labelled two-dimensional sentiment information for sentence, where one of axes is the inciting degree of the input sentence.
To utilize the inciting prediction capability, we fine-tune a pre-trained Chinese-BERT-base on the CVAT dataset and predict the incitation of the given news titles.

% Justin
\paragraph{Bias of the title} reflects journalist's prejudice against the person or organization in a news headline. Given a pair of a title and an name entity in the title, three domain experts are asked to annotate whether the journalists' stance toward the entity is positive, neutral or negative by their wording. We model the task as aspect-level sentiment analysis with fine-tuned Chinese RoBERTa~\cite{liu2019roberta}. Specifically, we encoded the news title using RoBERTa and extract hidden representations of the name entity as feature vectors, which is then aggregated with a mean pooling layer, followed by a three-way MLP classifier. To calculate the bias degree, we average the scores of non-neutral cases over all name entities in the headline.

Due to the scarcity of labeled data, we regularized the model with adversarial training~\cite{miyato2016adversarial} and apply weak supervision over the unlabeled data with pseudo labels of ensemble models. Moreover, we search the task-specific representations in RoBERTa layer-by-layer, to keep the model lightweight-yet-efficient during the inference phase.

\begin{table}[t]
\centering
\begin{tabular}{@{}lll@{}}
\toprule
Model             & Dev & Test \\
\midrule
RoBERTa           & 0.4954 & 0.5353  \\
\midrule
+ optimal layer searching           & 0.5963 & 0.6379 \\
+ adversarial training              & 0.5255 & 0.5742 \\
+ both                              & 0.6454 & 0.6539 \\
\midrule
weak supervision. & 0.6744 & 0.6922 \\
\bottomrule
\end{tabular}
\caption{The performance report of proposed aspect-level sentiment classifier (F1 score).}
\label{table:stance_classifier_performance}
\end{table}
The performance evaluation with the ablation study of our model is reported in Table~\ref{table:stance_classifier_performance}. We use macro F1-score as our evaluation metric for tackling data imbalance. We can observe that all the proposed methods clearly improve the performances, especially for optimal layer searching, which not only reduces the model complexity, but also significantly improve the test F1 score by $0.1026$.

% ZY
\paragraph{Subjectivity of the content} is defined as the portion of subjective words in a news. We utilize a domain expert to label 26 piece of news contents. The expert is asked to select arbitrary text span and decide whether it is a quotation, objective statement, or a subjective statement. We formulate this dataset as a token classification task. A Chinese-BERT-base pretrained model is finetuned on this dataset. During inference, we first classify if a sentence is a quotation or not, then we only classify subjective and objective sentences from non-quotation sentences. This two-stage procedure is performed because we are only interested in the subjective statement made by the journalist, and we shall neglect those from respondents. In the end, we use the number of token classified as subjective divided by the number of token in the whole news as the indicator of subjectivity of a piece of news.

\begin{figure*}[h!]
\centering
\includegraphics[width=\linewidth]{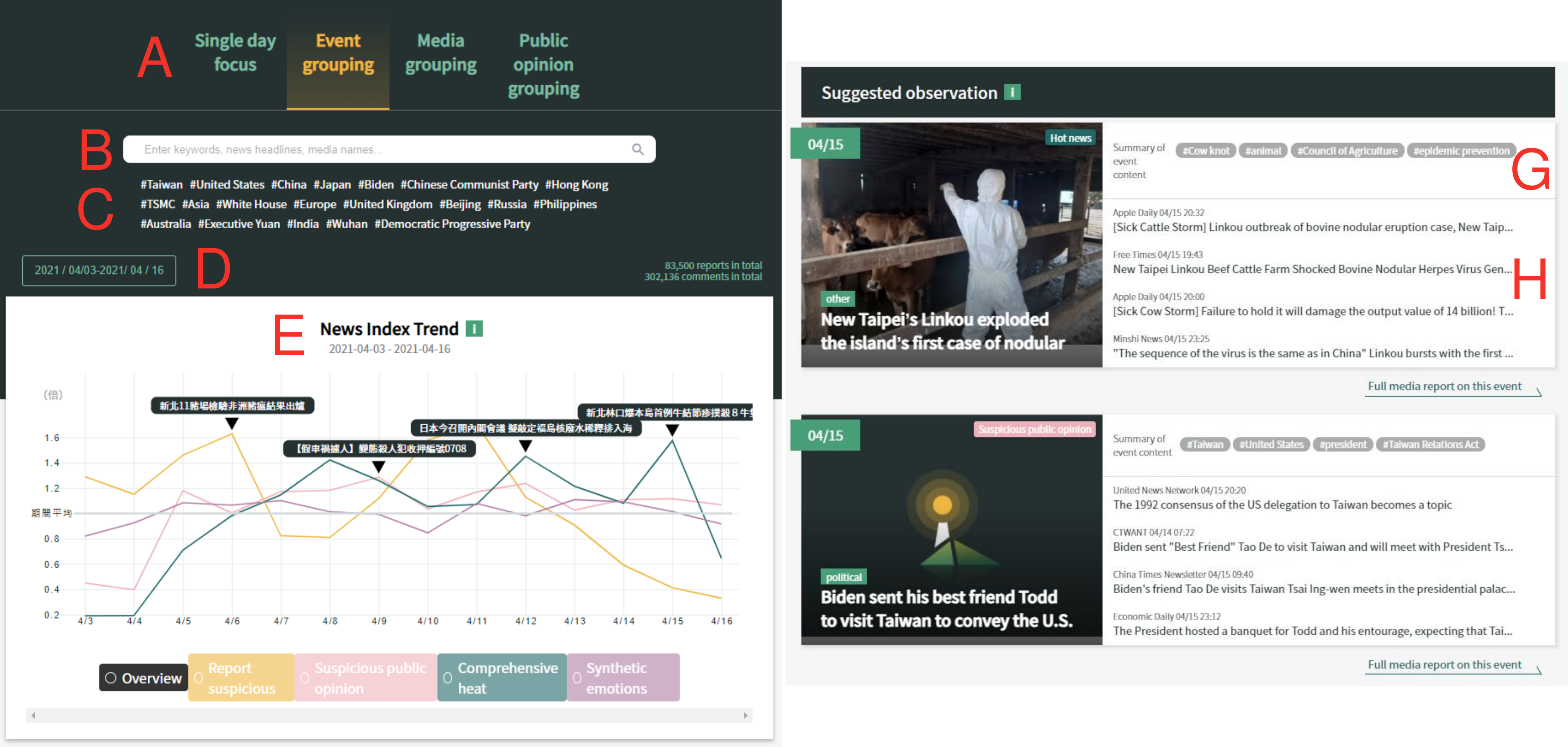}
\caption{The user interface of \Islander (translated by Google Translation). The components are: (a) selection of grouping by events, media, or public opinion, (b) keyword search, (c) trending keywords, (d) time span selection, (e) a chart of news metrics over time, (f) an event, with the selected event title, (g) automatically generated tags, (h) articles belonging to the event.}
\label{fig:interface}
\end{figure*}

\subsection{Suspicion Level of Public Opinions}

Besides mainstream medium, \Islander also models the public opinions regarding detected news events. Here we illustrate how our system can detect the coordinated behaviors and model public reaction to news events.

\paragraph{Coordinated behaviors} can reflect whether the public opinions come from normal users.
Considering that the opinions may be manipulated, our system focuses on detecting unnatural behaviors and then alarms the readers about its trustworthiness.
To check if there is coordinated behavior that affects the collected social opinions, we learn user embeddings using the co-occurrence relations. In particular, for $user_i$ and $user_j$, we construct their contingency table which is composed of 4 factors:
\begin{enumerate}
    \item frequency of $user_i$ and $user_j$ both occur in an article.
    \item frequency of $user_i$ occurs in an article, but $user_j$ does not.
    \item frequency of $user_j$ occurs in an article, but $user_i$ does not.
    \item frequency of $user_i$ and $user_j$ both do no occur in an article.
\end{enumerate}

Given the table, we measure the association of $user_i$ and $user_j$ using the phi coefficient. The association values are evaluated for all user pairs. Based on the association values, we construct a user network where the nodes are users and $node_i$ and $node_j$ would exist an edge if the association value of $user_i$ and $user_j$ is larger than the pre-defined threshold. We then apply node2vec~\cite{grover2016node2vec} on the user network to learn the distributional representation for each user. With the user embeddings, we can efficiently query the global similar users by K-nearest neighbor search. Furthermore, we can detect the coordinated behaviors by dynamically clustering the users who reply to the same event and check if there is a dense community.

\paragraph{Bot-style writing} may not be considered as the part of public opinions.
%Similar as metrics for news reports, we apply the well-tuned BERT models to predict the sentiment and inciting score for each comment. Besides, w
Hence, we find the frequent phrase in comments and calculate the occurrence ratio to detect whether bots are involved in the discussion.

% Justin 
\subsection{Popularity Level}

We proposed two indicators to measure the popularity of an event in the media and public opinions. The first indicator is the size of the detected cluster, which shows the amount of news regarding to an event. The second indicator is the amount of public opinion related to the events. Specifically, we periodically crawled the social discussions from LineToday and PTT, a mainstream chat forum in Taiwan, and we would check if a discussion thread is related to a detected event by comparing Levenshtein distance between the news title and the post title. The amount of relevant discussions would be added up as the popularity of the event in the social community.

% ZY
\subsection{Sentiment Level}
We construct our sentiment dataset from various sources, including posts and comments in a Taiwanese forum, post from a Chinese social media, and post from Twitter. Again, a Chinese-based BERT is fine-tuned on these datasets to produce the sentiment scores for both news content and user comments.

%\subsection{Metrics for Public Opinion}
% Justin.
% - overview.
% - coordinate behaviors.
% - stats features (incite, ).
% - meta information.

%\subsection{Meta-Information}

%We monitor the meta-information of discussions, such as the amount of involved 

%\Islander also monitor meta-information of 

\section{User Interface}
The web interface of {\Islander} is shown in Figure~\ref{fig:interface}.
The users can select whether they want the news to be grouped by events, by medium, or by public opinions (part A).
We provide a keyword search function and time span selection for the users to choose what they are interested in (part B and C).
A chart of news metrics over time is shown so that we can easily look for unusual events (part E).
Events are listed with their title, category and image (part F), along with tags generated from our tag generation module and all relevant articles (part G and H).

\section{Conclusion}
We present the {\Islander} system for real-time online news monitoring and analysis.
Several metrics are proposed as news quality measurement, and we leverage state-of-the-art technologies to develop models for measuring the metrics.
A web interface is integrated which lets user to browse events, select specific time span, search news by keywords, and analyze the events and media outlets.
We make the system publicly available and it is updated automatically on a regular basis.
We hope that {\Islander} could aid users to consume large volume of online news and make them aware of the subjectivity and possible manipulation.
In the future, we will constantly improve our system. The modules in {\Islander} can be further improved with new advances in natural language processing technologies or new data sources.

\bibliography{anthology,custom}
\bibliographystyle{acl_natbib}

\end{document}